\documentclass[lettersize,journal]{IEEEtran}
\usepackage{hyperref}
\usepackage{enumitem}
\usepackage{amsmath,amssymb,amsfonts}
\usepackage{booktabs, multirow} 
\usepackage{soul}
\usepackage[table]{xcolor} 
\usepackage{changepage,threeparttable} 
\usepackage{caption}
\usepackage{subcaption}
\usepackage{adjustbox}
\usepackage{epstopdf}
\usepackage{amsmath}
\usepackage{algorithm,algcompatible}
\usepackage{algpseudocode}
\usepackage{tikz}
\usepackage{pgfplots}
\usepackage{pgfplotstable}
\pgfplotsset{compat=1.7}
\usetikzlibrary{arrows.meta,
                chains,
                positioning}
\newtheorem{definition}{Definition}                
\usepackage{amsmath}

\begin{document}

\title{Graph-enabled Reinforcement Learning for Time Series Forecasting with Adaptive Intelligence}

\author{Thanveer Shaik,  Xiaohui Tao, Haoran Xie, Lin Li, Jianming Yong, and Yuefeng Li
\thanks{Thanveer Shaik and Xiaohui Tao are with 
the School of Mathematics, Physics \& Computing, University of Southern Queensland, Toowoomba, Queensland, Australia (e-mail: Thanveer.Shaik@usq.edu.au, Xiaohui.Tao@usq.edu.au).}
\thanks{Haoran Xie is with the Department of Computing and Decision Sciences, Lingnan University, Tuen Mun, Hong Kong (e-mail: hrxie@ln.edu.hk)}
\thanks{Lin Li is with the School of Computer and Artificial Intelligence, Wuhan University of Technology, China (e-mail: cathylilin@whut.edu.cn)}
\thanks{Jianming Yong is with the School of Business at the University of Southern Queensland, Queensland, Australia (e-mail: 	Jianming.Yong@usq.edu.au)}
\thanks{Yuefeng Li is with the School of Computer Science, Queensland University of Technology, Brisbane, Australia (e-mail: y2.li@qut.edu.au).}
}



\maketitle

\begin{abstract}
Reinforcement learning is well known for its ability to model sequential tasks and learn latent data patterns adaptively. Deep learning models have been widely explored and adopted in regression and classification tasks. However, deep learning has its limitations such as the assumption of equally spaced and ordered data, and the lack of ability to incorporate graph structure in terms of time-series prediction. Graphical neural network (GNN) has the ability to overcome these challenges and capture the temporal dependencies in time-series data. In this study, we propose a novel approach for predicting time-series data using GNN and monitoring with Reinforcement Learning (RL). GNNs are able to explicitly incorporate the graph structure of the data into the model, allowing them to capture temporal dependencies in a more natural way. This approach allows for more accurate predictions in complex temporal structures, such as those found in healthcare, traffic and weather forecasting. We also fine-tune our GraphRL model using a Bayesian optimisation technique to further improve performance. The proposed framework outperforms the baseline models in time-series forecasting and monitoring. The contributions of this study include the introduction of a novel GraphRL framework for time-series prediction and the demonstration of the effectiveness of GNNs in comparison to traditional deep learning models such as RNNs and LSTMs. Overall, this study demonstrates the potential of GraphRL in providing accurate and efficient predictions in dynamic RL environments.
\end{abstract}


\section{Introduction}
The emergence of Machine Learning (ML) in healthcare signifies a paradigm shift towards automating clinician tasks and augmenting patient care capabilities~\cite{Zhang2022}. Amidst the evolving ML landscape, Federated Learning has gained traction for preserving data privacy while constructing sophisticated server models~\cite{shaik2022fedstack}. Reinforcement Learning (RL), another ML strategy, has demonstrated substantial improvements in prediction performance and decision-making tasks~\cite{gaoreinforcement, forman2019can}. RL's application is particularly noteworthy in controlling autonomous systems, such as robots and drones, training them to make optimal decisions in real-time based on environmental sensor data.

In various sectors, including healthcare, traffic, and weather forecasting, Early Warning Systems (EWS) play a pivotal role. They analyze real-time monitoring data and issue alerts for potential issues, facilitating proactive responses. RL-based EWS can adapt over time, refining their predictions and supporting clinical decision-making. This adaptability has proven effective in applications like predicting hospital readmissions and sepsis detection.

\begin{figure}
    \centering
    \includegraphics[width=\columnwidth]{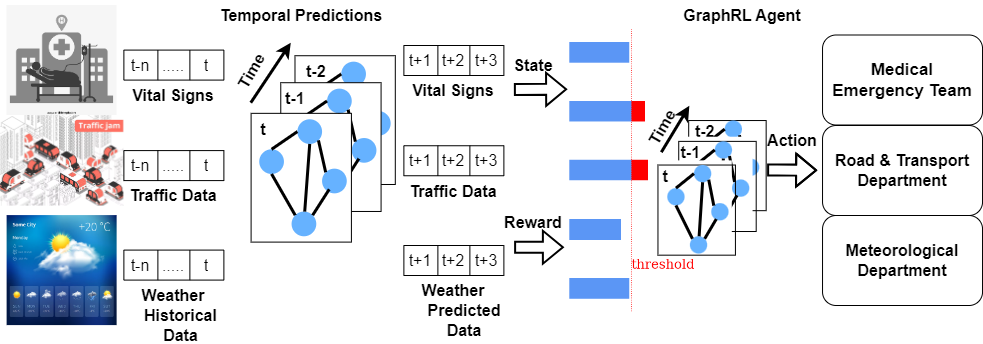}
    \caption{Graphical Abstract}
    \label{fig:graph_abs}
\end{figure}

Time-series data modeling, vital in monitoring and predicting future states, has seen advancements with deep learning models like Recurrent Neural Networks (RNNs) and Long Short-Term Memory (LSTM) networks~\cite{wang2022reinforcement}. These models are adept at capturing temporal dependencies, yet face limitations in handling irregularly structured data and integrating complex graph structures. This study introduces the GraphRL framework, an innovative amalgamation of RL and Temporal Graphical Convolutional Networks (T-GCN), aiming to surpass the constraints of traditional deep learning models in time-series prediction. GraphRL's design facilitates handling complex temporal structures and incorporates additional information such as node and edge attributes, as depicted in Fig.~\ref{fig:graph_abs}. The GraphRL framework's core contributions include:
\begin{itemize}
\item A versatile framework capable of providing early warnings and monitoring in complex settings.
\item A customizable RL environment designed for effective forecasting in dynamic domains like healthcare and traffic systems.
\item A novel approach to virtual monitoring of predicted states in RL, enhancing decision-making and intervention capabilities.
\end{itemize}

Our comparative analysis with state-of-the-art models across various datasets showcases GraphRL's superior performance, underscoring its potential as a versatile solution for time-series prediction challenges.

The paper is organized as follows: Section~\ref{relatedworks} reviews existing literature on self-learning systems and prediction tasks. Section~\ref{problem} outlines the research problem. The proposed GraphRL framework and its algorithm are detailed in Section~\ref{method}. Section~\ref{experiment} describes the datasets and baseline models used for evaluation. The performance of the predictive RL environment and agent is compared with baseline models in Section~\ref{results}. Section~\ref{optimization} discusses the fine-tuning of the framework's hyperparameters using Bayesian optimization. The paper concludes in Section~\ref{conclusion}.

\section{Related Works}\label{relatedworks}
\subsection{Self Learning Systems}
Self-learning systems, particularly those utilizing Reinforcement Learning (RL), have seen significant advancements in various applications. For instance, Shin et al.~\cite{shin2022modeling} introduced a dual-agent framework in mobile health, effectively demonstrating user modeling and behavior intervention strategies. This work underscores the potential of RL in personalizing user experiences, a concept that aligns with our GraphRL framework's goal of adaptive learning in dynamic environments. Similarly, Taylor et al.~\cite{taylor2021awareness} applied RL in modeling maladaptive eating behaviors, further showcasing RL's versatility in behavior prediction and modification. Chen et al.~\cite{chen2022mirror} developed the MIRROR framework, emphasizing the rapid learning capabilities of RL in human behavior modeling. These advancements set a precedent for our work in complex sequential decision-making tasks. Zhou et al.'s~\cite{zhou2018personalizing} CalFit app and Li et al.'s~\cite{li2021hierarchical} method in autonomous driving highlight RL's efficacy in personalized goal setting and complex urban scenario navigation, respectively, which are foundational to our GraphRL framework's approach in handling dynamic and intricate patterns in data.

\subsection{Early Detection of Patient Deterioration}
In the healthcare domain, early detection of patient deterioration is vital. Traditional vital signs monitoring, as discussed by Asiimwe et al.~\cite{asiimwe2020vital} and evaluated by Scully et al.~\cite{scully2017evaluating}, Baig et al.~\cite{baig2016machine}, and others, has laid the groundwork for our study. These works highlight the importance of continuous monitoring and early warning systems (EWS), which are integral to GraphRL's objective. The limitations in existing methods, such as the need for manual calculations and the inability to handle large, unstructured data effectively, are addressed in our framework through the integration of GNNs, which can process complex temporal data more efficiently.

\subsection{Vital Signs Prediction}
Vital signs prediction has been explored through various machine learning models. Alghatani et al.~\cite{alghatani2021predicting} and Youssef et al.~\cite{youssef2021vital} demonstrated the use of traditional machine learning in mortality prediction and vital signs forecasting, respectively. Harerimana et al.'s~\cite{harerimana2022multi} work with multi-head transformers and Xie et al.'s~\cite{xie2022deepvs} DeepVS model highlight the potential of deep learning in this domain. However, these methods often assume equally spaced and ordered data and lack the ability to incorporate complex graph structures, limitations our GraphRL framework aims to overcome.

\subsection{Temporal Graphical Convolutional Networks (T-GCN)}
The integration of T-GCN within our GraphRL framework is pivotal. T-GCNs, known for their ability to capture temporal dependencies and complex relationships in graph-structured data, offer significant enhancements in processing time-series data~\cite{elsir2023hlgst}. This technology addresses limitations in traditional deep learning models by effectively managing irregular time intervals and integrating additional contextual information (such as node and edge attributes) for richer data representation~\cite{wang2020deep}. The inclusion of T-GCN in GraphRL allows for a more nuanced understanding and prediction of dynamic systems~\cite{chen2024knowledge}, making it highly suitable for applications in healthcare monitoring, traffic forecasting, and weather prediction. The capability of T-GCNs to handle non-linear and complex temporal patterns~\cite{huttel2023mind} aligns with the core objectives of GraphRL, pushing the boundaries of current self-learning systems in real-world scenarios.

In summary, while existing works in self-learning systems, patient deterioration detection, and vital signs prediction have laid a strong foundation, our GraphRL framework aims to address their limitations by introducing a novel approach that combines the strengths of GNNs and RL. This approach allows for a more sophisticated handling of temporal dependencies and real-time monitoring, which is crucial in dynamic environments such as healthcare, traffic management, and weather forecasting. 

The motivation behind the use of RL in our framework primarily arises from the need to tackle the challenges of multi-step time series prediction, where traditional supervised learning approaches may encounter limitations. Although supervised learning methods like GNN+Bert and GNN+TCN are indeed common and effective for time series forecasting, RL offers a unique advantage in dealing with situations where errors can accumulate over time, especially in dynamic environments. RL enables our predictive GraphRL Environment not only to forecast future states but also to actively influence decision-making, a capability particularly valuable in applications such as healthcare monitoring and the gaming industry.

\section{Research Problem}\label{problem}
The research problem addresses deep learning challenges in predicting future states of a complex and dynamic Reinforcement Learning (RL) environment and adaptively learning latent behavior patterns of data.

\begin{definition}[Vital Parameters and Time-Series Forecasting]
In the context of our framework, we consider a set \(V\) of \(n\) vital parameters, denoted as \(V_{t} = \{v^1, v^2, \ldots, v^n\}\), which represent continuous time-series data reflecting the health status of a subject \(S\). These vital parameters are dynamic and change over time, providing valuable insights into the subject's well-being. To facilitate time-series forecasting, we segment these continuous vital parameters into time windows, denoted as \(T\), which encompasses data points from the past (\(V_{t-2}, V_{t-1}, V_{t}\)) and extends into the future (\(V_{t+1}, V_{t+2}, \ldots, V_{t+n}\)). Non-linear models are trained on historical data within the time windows \(\{V_{t-2}, V_{t-1}, V_{t}\}\) to predict future values \(\{V_{t+1}, V_{t+2}, \ldots, V_{t+n}\}\).
\end{definition}

\begin{definition}[Learning Agents and Markov Decision Process]
Following the training phase, subject \(S\) is associated with a group of learning agents that operate based on the principles of the Markov Decision Process (MDP). This MDP is a 5-tuple denoted as \(M = (S, A, P, R, \gamma)\), and it forms the foundation for continuous monitoring and pattern learning of the vital parameters \(V_{t} = \{v^1, v^2, \ldots, v^n\}\). Here's a breakdown of each component:
\begin{itemize}
    \item \(S\) represents a finite state space, where \(s_{t} \in S\) signifies the state of an agent at a specific time \(t\),
    \item \(A\) is the set of actions available to each agent, and \(a_{t} \in A\) represents the action taken by the agent at time \(t\),
    \item \(P\) is a Markovian transition function \(P(s, a, s')\) that quantifies the probability of the agent transitioning from state \(s\) to state \(s'\) while executing action \(a\),
    \item \(R\) is a reward function \(R: S \times A \rightarrow \mathbb{R}\) that provides an immediate reward \(R(s, a)\) for the action \(a\) performed in state \(s\),
    \item \(\gamma\) is a discount factor, ranging between 0 and 1, which emphasizes immediate rewards over future rewards.
\end{itemize}
\end{definition}

\begin{equation}\label{reward}
    R(s_{t}, a_{t}) = \sum_{t=0}^{\infty} \gamma^{t} r_{t},
\end{equation}
This equation returns the immediate reward \(R(s, a)\) for the action taken in state \(s\), as defined in Eq.~\ref{reward}.

\begin{figure*}[!h]
    \centering
    \includegraphics[width=0.9\textwidth]{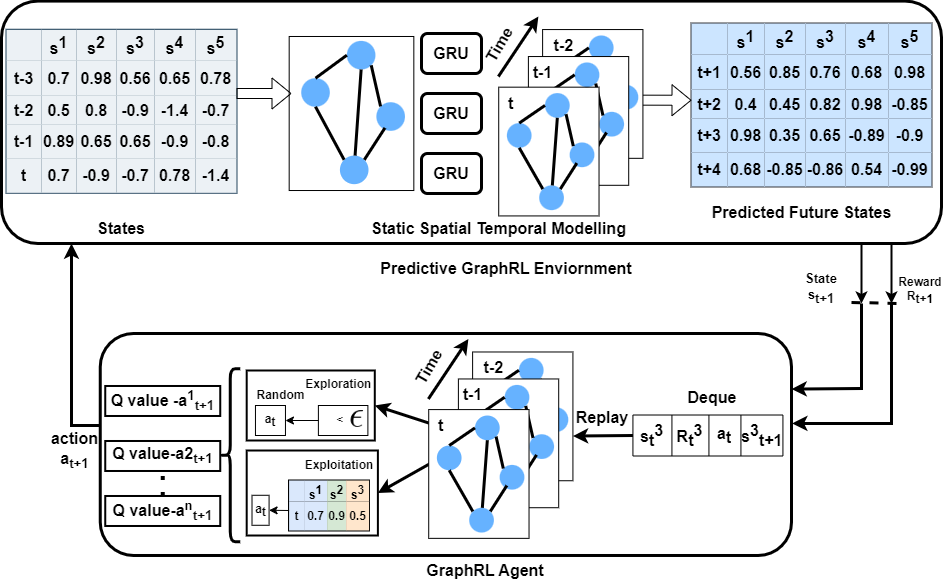}
    \caption{GraphRL Framework}
    \label{fig:methodology}
\end{figure*}

\section{GraphRL Framework}\label{method}
To address the research problem, a novel graphical neural network (GNN) enabled reinforcement learning (RL) framework is proposed. In the graph-enabled RL framework(GraphRL), two GNNs are deployed: one for forecasting time-series data and another for Q-function approximation as shown in Fig.~\ref{fig:methodology}. The proposed framework is demonstrated in Fig.~\ref{fig:methodology} in which the interaction between an environment and an AI agent is illustrated. As discussed in the research problem, finite MDP is adopted to formulate the process of modelling current and past states. 
\subsection{Predictive GraphRL Environment}\label{Environment}
The primary objective of the proposed study is to learn from the past and current states of a dynamic environment and predict the future states of the complex environment. To achieve this objective, we propose a predictive monitoring environment which is responsible for defining the observation space with state $s_{t}^{i} \epsilon S$ where $i=0,1,2,...n$, action space with actions $a_{t}^{j} \epsilon A$ where $j=1,2,3,...m$, and rewards $R$ for each action taken by the agent as it transitions from a state $s_{t}$ to $s_{t+1}$ in a real-world scenario. For example, consider a subject in a dynamic environment whose current state is denoted as $V_{t}={v^1,v^2,...,v^n}$ at time $t$. Similarly, the subject holds historical data of their state at times $t-1, t-2, t-3,...,t-n$. In traditional reinforcement learning formulations, the monitoring environment is a static entity that cannot forecast future states, which might affect the subjects in the environment.

\subsubsection{T-GCN Forecast}
Forecasting the future states of a subject before a few time steps can revolutionise the most dynamic industries such as gaming, healthcare, and so on by identifying the deteriorating state of the subject in the environment. To predict the future state in a reinforcement learning environment, a temporal graph convolutional network(T-GCN) is adopted. The graphical network is trained with past and current states at their timestamps in a supervised approach as shown in Eq.~\ref{pred-eq}. The training process also includes the features leading to those states.  

\begin{equation}\label{neuron} 
   y=f(b + \sum _{i=1}^{n} v_{i}.w_{i})
\end{equation}

\begin{equation}\label{pred-eq}
\begin{split}
y(v) = \sum_{i=1}^{n} Activation1(b+w_{i} v_{i})\\
y(v) = Activation1(\frac{e^{y_{i}}}{\sum _{j=1}^{k} e^{y_{j}}})
\end{split}
\end{equation}

Eq.~\ref{neuron} describes a basic neural network neuron, with \( y \) as the output, \( f \) as the activation function, \( b \) as the bias, \( v_i \) as the input features, and \( w_i \) as the weights; while Eq.~\ref{pred-eq} involves an 'Activation1' function, computing a weighted sum of inputs and normalizing these outputs, possibly into a probability distribution, akin to a softmax function.

\subsubsection{Static Spatial-temporal modelling}
A two-layered graphical network is adopted for Spatial-temporal modelling, a spatial modelling layer is based on a graphical convolutional network (GCN), and a temporal layer based on recurrent neural networks (RNN) is configured. The spatial layer is responsible to capture spatial features among nodes which are input states $s_{t}^{i} \epsilon S$. This can be achieved by constructing Fourier transform filter and it acts on the graph nodes and its first-order neighbourhood. In this study, a static graph temporal signal is adopted in which the node positions in the graph remain the same and the label information is dynamic. The spatial layer is to set the static graph with nodes as input states $s_{t}^{i} \epsilon S$. The two-layered GCN model is defined in Eq.~\ref{spatial}.

\begin{equation}\label{spatial}
    f(X,A)=\sigma (\hat{A} Relu(\tilde{A} X W_{0})W_{1})
\end{equation}

Where X is the input matrix, A represents the graph adjacency matrix, $\hat{A}$ and $\Tilde{A}$ represent the preprocessing step and self-connection structure respectively. $W_{0}, W_{1}$ represents weights of the first and second layers of ST-GCN, and $\sigma(\cdot)$. $Relu()$ is an activation function. 

Temporal modelling is based on the RNN variant gated recurrent unit (GRU)~\cite{cho2014properties} which has a simple structure and faster training ability. In the GRU model, an update gate $z_{t}$ controls the degree of information retrieved from the previous state and a reset gate $r_{t}$ controls the degree of ignoring the status information at the previous moment are configured as shown in Eq.~\ref{temporal}.

\begin{equation}\label{temporal}
   {\displaystyle {\begin{aligned}z_{t}&=\sigma _{g}(W_{z}[f(X_{t},A),h_{t-1}]+b_{z})\\r_{t}&=\sigma _{g}(W_{r}[f(X_{t},A),h_{t-1}]+b_{r})\\
   {\hat {h}}_{t}&=\phi _{h}(W_{h}[f(X_{t},A),h_{t-1}](r_{t}\odot h_{t-1})+b_{h})\\h_{t}&=z_{t}\odot h_{t-1}+(1-z_{t})\odot {\hat {h}}_{t}\end{aligned}}} 
\end{equation}

${\displaystyle x_{t}}:$ input vector,
${\displaystyle h_{t}}:$ output vector,
${\displaystyle {\hat {h}}_{t}}:$ candidate activation vector,
${\displaystyle z_{t}}:$ update gate vector,
${\displaystyle r_{t}}:$ reset gate vector,
${\displaystyle W}, {\displaystyle U}, {\displaystyle b}:$ parameter matrices and vector, 
$\sigma _{g}:$ The original is a sigmoid function,
${\displaystyle \phi _{h}}$ The original is a hyperbolic tangent.

In the training process, the T-GCN model predicts the future states at time $t+1, t+2, t+3,...,t+n$ as $\hat{Y_{t}}$ and compares with the real-data ${Y_{t}}$. This determines the loss function of the graph network~\cite{yu2017spatio} as shown in Eq.~\ref{loss}. To avoid over-fitting problems in the training process, the loss function is optimised with L2 regularisation $L_{reg}$ and a hyperparameter $\lambda$.

\begin{equation}\label{loss}
    loss = ||\hat{Y_{t}} - {Y_{t}}||+ \lambda L_{reg}
\end{equation}

\subsection{Predictive GraphRL Environment Algorithm}

\begin{algorithm}[!ht]
\scriptsize
\caption{Predictive GraphRL Environment} \label{alg1:cap}
\begin{algorithmic}[1]
\Ensure{\textbf{Input:} time series data $\mathcal{D}=\{s_{t-n},\ldots,s_{t-2},s_{t-1},s_{t}\}$; a set of labels $\mathcal{K}=\{1,2,\dots,K\}$}
\Ensure{\textbf{Output:} Predicted time series data of $\mathcal{K}$, a set of labels, in the form of states $\{s_{t+1},s_{t+2},s_{t+3},s_{t+4}\}$}
\State Define $forecast\_model \leftarrow T-GCN Model$
\State $Train(forecast\_model) \leftarrow forecast\_model(\mathcal{D})$
\State $\{s_{t+1},s_{t+2},s_{t+3},s_{t+4}\} \leftarrow forecast\_model(predict)$
\State $Initialization: observation\_space=\{s_{t}^{i} \in S\}, action\_space=\{a_{t}\in A\}, reward R$
\State $Set\ monitor\_length= N$
\If{$action$ is appropriate}
\State $R \leftarrow +reward$
\Else
\State $R \leftarrow -reward$
\EndIf
\State $monitor\_length \leftarrow N-1$
\State $s_{t+1} \leftarrow s_{t}(monitor\_length)$
\If {$N=0$}
\State $done \leftarrow True$
\Else
\State $done \leftarrow False$
\EndIf
\State $visualize(a_{t}, R, vital\ signs)$
\State $initial\_state \leftarrow s_{t}[0]$ \Comment{reset environment}
\end{algorithmic}
\end{algorithm}

Algorithm~\ref{alg1:cap} outlines the creation of the Predictive GraphRL Environment, a crucial component of the proposed GraphRL framework. This environment leverages the T-GCN, chosen for its effectiveness in capturing spatial-temporal dynamics essential for complex systems. The T-GCN's ability to forecast future states, in addition to analyzing current and historical data, makes it invaluable for critical applications like health monitoring and traffic management, where early detection and timely response are key. A significant feature of this algorithm is its reward mechanism, which is instrumental in guiding the learning process. By awarding rewards based on the suitability of the agent's actions, the environment ensures that the agent's policy is aligned with the primary objectives of accurate forecasting and effective intervention. This design was motivated by the need for a proactive system, capable of not only forecasting but also informing real-time decision-making. The algorithm is meticulously structured to set up the observation space, action space, and reward policy based on predicted states. The initial lines (1-3) justify the use of the T-GCN model, especially for its applicability in dynamic and nonlinear data contexts like vital sign monitoring. The subsequent lines (4-5) are dedicated to initializing the environment, forming the basis for RL-driven decision-making. The reward policy, detailed in lines 6-10, aligns with standard RL practices, promoting actions that yield beneficial outcomes. Finally, lines 11-19 focus on continuous monitoring and adaptation, a critical aspect for applications that demand real-time responsiveness, such as in healthcare scenarios. This algorithm represents a significant step in advancing the field of RL, moving from passive observation to an active role in shaping decisions.

\subsection{GraphRL Agent}

In this study, the Deep Q-Networks (DQN) algorithm is used. The Deep Q-Networks (DQN) algorithm, developed by Google's DeepMind, was initially designed for playing Atari games. This algorithm enabled the AI to learn game strategies directly from visual input, without requiring pre-programmed rules or prior game-specific training. In this algorithm, the Q-Learning functions are approximated using the proposed T-GCN model, and the learning agent is rewarded based on the graph network prediction of the right action for the current state. 

\subsubsection{Q-Function Approximation}
T-GCN model used in this study to approximate the Q-Function for each action in the action space as shown in Fig.~\ref{fig:methodology}. The model is configured with parameters such as the relu activation function, mean square error as loss function, and Adam optimiser. The model gets trained with the state and its corresponding action. The learning agent performs an action $a_{t} \epsilon A$ for a transition from state $s_{t}$ to $s_{t}^{'}$ and achieves a reward $R$ for the action. In this transition process, the maximum of the Q-function in Eq.~\ref{modified_action_value} is calculated, and the discount of the calculated value uses a discount factor $\gamma$ to suppress future rewards and focus on immediate rewards. The discounted future reward is added to the current reward to get the target value. The difference between the current prediction from the neural networks and the calculated target value provides a loss function. The loss function is a deviation of the predicted value from the target value and it can be estimated from Eq.~\ref{loss function}. The square of the loss function allows for the punishment of the agent for a large loss value. 

\begin{equation}\label{modified_action_value}
Q^{\pi}(s,a)  = E_{\pi}\biggl\{\ \sum _{t=0}^{\infty} \gamma^{t}R(s_{t},a_{t},\pi(s_{t}))| s_{0}=s, a_{0}=a \biggl\}
\end{equation}

\begin{equation}\label{loss function}
   \displaystyle{ loss = (\underbrace{R+\gamma \cdot max(Q^{\pi^{*}}(s,a))}_{target\_value}- \underbrace{Q^{\pi}(s,a)}_{predicted\_value})^{2}}
\end{equation}

\subsubsection{Exploration and Exploitation}
The concepts of exploration and exploitation are at odds with each other. Exploration involves randomly selecting actions that have not been performed before to uncover more possibilities and enhance the agent's understanding. Exploitation, on the other hand, entails selecting actions based on past experiences and knowledge to maximize rewards. To balance the trade-off between exploration and exploitation, different strategies such as the greedy algorithm, epsilon-greedy algorithm, optimistic initialization, and decaying epsilon-greedy algorithm are employed. This study proposes controlling the exploration rate by multiplying the decay by the exploration rate. This approach reduces the number of explorations during execution as the agent learns patterns and maximizes rewards to achieve high scores. As the T-GCN model is retrained with previous experiences in the replay, the decay is multiplied by the exploration rate depending on the agent's ability to predict the right actions. All these parameters are defined as hyper-parameters for DQN learning agents.

\subsubsection{GraphRL Agent Algorithm}
\begin{algorithm}[!ht]
\scriptsize
\caption{Learning Agent}\label{alg2:cap}
\begin{algorithmic}[1]
\State Initialize $\gamma, \epsilon, \epsilon_{decay}, \epsilon_{min}, memory=\emptyset, batch\_size$ 
\State Define $model \leftarrow T-GCN\_model$
\State $memory \leftarrow append(s_{t}, a_{t}, R, s_{t+1})$
\If {$np.random.rand() < \epsilon$} \Comment{Exploration}
\State $action\_value \gets random(a_{t})$
\Else \Comment{Exploitation}
\State $action\_value \gets model.predict(s_{t})$
\EndIf
\State $minibatch \gets random(memory, batch\_size)$
\For {$s_{t}, a_{t}, R, s_{t+1}, done$ in $minibatch$}
\State $target \gets R$
\If {not done}
\State $target \gets R + \gamma \cdot \max(model.predict(s_{t+1}))$
\EndIf
\State $target\_f \gets model.predict(s_{t})$
\State $target\_f[a_{t}] \gets target$
\State $model.fit(s_{t}, target\_f)$
\EndFor
\If {$\epsilon \geq \epsilon_{min}$}
\State $\epsilon \gets \epsilon \cdot \epsilon_{decay}$
\EndIf
\end{algorithmic}
\end{algorithm}

Algorithm~\ref{alg2:cap} introduces the GraphRL Agent, presents the functionality of the GraphRL Agent within a complex action-state environment, utilizing T-GCN for Q-function approximation. This integration enables effective handling of spatial-temporal data complexities, enhancing decision-making. The algorithm's design is founded on a strategic balance between exploration and exploitation, achieved via an epsilon-greedy strategy, crucial for adaptive learning and continual improvement in decision-making. It starts with initializing key parameters (Line 1), defining a T-GCN model for handling complex data structures (Line 2), and storing memories for experience replay (Line 3). The agent's learning process involves iterative learning from minibatches of experiences, computing target Q-values, and adjusting its policy (Lines 8-21), with a dynamic adjustment of the exploration rate (Lines 16-18). The predict() function, pivotal in the exploitation phase, utilizes the T-GCN model's predictions to guide actions, showcasing the algorithm's advanced approach in navigating dynamic environments through a blend of exploration and strategic exploitation.

\subsubsection{Implementation Algorithm}

\begin{algorithm}[ht]
\scriptsize
\caption{Proposed GraphRL Framework Implementation}\label{alg3:cap}
\begin{algorithmic}[1]
\Require{\textbf{Input:}
\Statex $\mathcal{C}=\{1,2,\ldots,C\}$: set of subjects
\Statex $\mathcal{V}=\{1,2,\ldots,V\}$: set of vital signs
\Statex $\mathcal{M}=\{1,2,\ldots,M\}$: number of episodes
}
\Ensure{\textbf{Output:} Rewards achieved by Agents in each episode.}
\State $env \leftarrow ForecastingEnvironment()$ \Comment{Algorithm~\ref{alg1:cap}}
\State $agent \leftarrow LearningAgent()$ \Comment{Algorithm~\ref{alg2:cap}}
\For {episode $m \in \mathcal{M}$}
\State $state \leftarrow env.reset()$
\State $score \leftarrow 0$
\For {time in range(timesteps)}
\State $a_{t} \leftarrow agent.action(s_{t})$
\State $s_{t+1}, R, done \leftarrow env.step(a_{t})$
\State $agent.memorize(s_{t}, a_{t}, R, s_{t+1})$
\State $s_{t} \leftarrow s_{t+1}$
\If {done}
\State $print(m, score)$
\State $break;$
\EndIf
\EndFor
\State $agent.replay(batch\_size)$
\EndFor
\end{algorithmic}
\end{algorithm}

Algorithm~\ref{alg3:cap} serves as the comprehensive implementation of the GraphRL framework, intricately combining the Predictive GraphRL Environment (Algorithm 1) with the GraphRL Agent (Algorithm 2). This pivotal algorithm orchestrates the real-time interactions between the agent and the environment, thus forming the operational core of the framework. It outlines the simulation scope, input parameters (subjects $\mathcal{C}$, vital signs $\mathcal{V}$, and episodes $\mathcal{M}$), and the output in the form of cumulative rewards. The initialization phase prepares the environment env and the agent for interaction. The episodic loop, encompassing the agent's action-response cycle, is vital for continuous learning and adaptation. Crucial to this process is the memorize() function, which stores experiences for later recall during experience replay, allowing the agent to learn from past actions and refine its decision-making strategy. This algorithm thus encapsulates the dynamic and iterative nature of the GraphRL framework, highlighting the importance of memory and experience in the realm of advanced reinforcement learning, and showcasing its functionality in complex, evolving environments.

\subsection{Bayesian Optimisation}\label{optimization}
Bayesian optimisation is a global optimisation method that uses a probabilistic model to guide the search for the optimal solution. The model is updated as new data points are sampled and evaluated, allowing the algorithm to improve its predictions over time by fine-tuning the $L_{reg}$, $\lambda$ and minimising the loss function defined in Eq.~\ref{loss}. The basic idea behind Bayesian optimisation is to model the objective function, f(x), as a Gaussian process (GP). The GP model is used to predict the objective function value at any point x, given the observations of the function at other points. The prediction is given by the posterior distribution of the GP, which is a Gaussian distribution with mean and variance given by Eq.~\ref{gp}.

\begin{equation}\label{gp}
 {\displaystyle \begin{aligned} \mu(x) &= k(x,X)^T(K + \sigma^2I)^{-1}y \\
\sigma^2(x) &= k(x,x) - k(x,X)^T(K + \sigma^2I)^{-1}k(X,x) \end{aligned} }
\end{equation}

Where X is the matrix of previously sampled points, y is the vector of corresponding function values, K is the Gram matrix of the covariance function evaluated at X, and $\sigma^{2}$ is the noise level in the function evaluations.
The next point to sample is chosen based on an acquisition function, which balances the trade-off between exploration and exploitation. Common acquisition functions include the probability of improvement and the expected improvement. Given a set of observed points (X,y) and a Gaussian process prior, Bayesian optimisation seeks the point x* that minimises the loss function value, given by Eq.~\ref{gp2}. The optimisation process continues iteratively, sampling new points and updating the GP model until a stopping criterion is met.

\begin{equation}\label{gp2}
 {\displaystyle EI(x) = \mathbf{E}[max(0,f(x)-f(x^*))]}
\end{equation}
Where x* is the current best point.

\section{Experiment}\label{experiment}
The primary objective of this study is to overcome deep learning challenges such as the assumption of equally spaced and ordered data~\cite{karagiorgi2022machine} and the lack of ability to incorporate graph structure where the data has a complex temporal structure~\cite{almasan2022deep}. These challenges are particularly relevant in domains such as health, weather, and traffic where it is important to analyze temporal patterns and make accurate forecasts for early warning systems. However, traditional deep learning models often fail to capture these complex patterns, limiting their effectiveness in these critical domains.

The proposed GraphRL framework is evaluated on three different forecasting applications: heart rate prediction, traffic forecast, and weather forecast. The framework predicts future events in the form of states and optimizes actions based on those predictions. The observation space is customized for each application and actions for the agent are pre-defined. The agent receives a reward for correctly predicting a state and communicating with the relevant team. The proposed approach is a generic framework that can be applied to monitor and predict time-series data and train an RL agent to learn the latent patterns of the monitoring process. The baseline models for comparison include traditional deep learning models such as GRU, LSTM, and RNNs.

\subsection{Datasets}
The GraphRL framework's testing with datasets from healthcare, traffic, and weather domains was a deliberate strategy to assess its versatility and robustness in handling diverse time-series data. The choice of these varied domains was intended to demonstrate the framework's adaptability and efficacy in different contexts. Each domain poses unique challenges: healthcare data's complexity and sensitivity, traffic data's dynamic patterns requiring real-time analysis, and weather data's intricate interplay of environmental factors. Successfully navigating these distinct datasets underscores the framework's capability for widespread real-world application. Additionally, using datasets from different fields facilitates a thorough evaluation of the framework, ensuring its versatility and effectiveness across various problem types and data structures. This comprehensive approach is vital for a tool designed for extensive applications in data analysis and prediction. Three datasets utilized for evaluating the GraphRL framework as shown in Fig.~\ref{tab:datasets}, each from a different domain: healthcare, traffic, and weather. In healthcare, the WESAD dataset, containing electrocardiogram (ECG) and photoplethysmogram (PPG) data from 17 participants, offers a rich source of biometric time-series data for pattern recognition analysis. The Los Angeles (LA) Traffic dataset, sourced from the Los Angeles Department of Transportation (LADOT), provides real-time urban traffic data like traffic counts and speeds, while the Large-Scale Traffic and Weather Events (LSTW) dataset, with data across the United States, uniquely combines traffic conditions and weather events, posing a multifaceted challenge for the framework.

\begin{table}[!ht]
\caption{Datasets}
\label{tab:datasets}
\scriptsize
\resizebox{\columnwidth}{!}{%
\begin{tabular}{lllll}
\hline
\multicolumn{1}{c}{\textbf{Dataset}} &
  \multicolumn{1}{c}{\textbf{Domain}} &
  \multicolumn{1}{c}{\textbf{Key Features}} &
  \multicolumn{1}{c}{\textbf{Statistics}} &
  \multicolumn{1}{c}{\textbf{Suitability}} \\ \hline
\begin{tabular}[c]{@{}l@{}}WESAD\\ \cite{schmidt2018introducing}\end{tabular} &
  Healthcare &
  \begin{tabular}[c]{@{}l@{}}Physiological data \\ (ECG, PPG),\\ motion data\\ (accelerometers)\end{tabular} &
  17 participants &
  \begin{tabular}[c]{@{}l@{}}Rich  biometric time-series\\ data, ideal for testing pattern\\ recognition in health-related\\ data.\end{tabular} \\ \hline
\begin{tabular}[c]{@{}l@{}}LA \\ Traffic\\ \cite{li2017diffusion}\end{tabular} &
  Traffic &
  \begin{tabular}[c]{@{}l@{}}Traffic counts, \\ speeds,\\ travel times\end{tabular} &
  \begin{tabular}[c]{@{}l@{}}Data from\\ LADOT\end{tabular} &
  \begin{tabular}[c]{@{}l@{}}Real-time urban traffic data,\\ useful for analyzing and\\ predicting dynamic traffic flows.\end{tabular} \\ \hline
\begin{tabular}[c]{@{}l@{}}LSTW\\ \cite{moosavi2019short}\end{tabular} &
  Weather &
  \begin{tabular}[c]{@{}l@{}}Traffic conditions,\\ weather events\end{tabular} &
  \begin{tabular}[c]{@{}l@{}}Data across\\ the United\\ States\end{tabular} &
  \begin{tabular}[c]{@{}l@{}}Combines traffic and weather\\ data, challenging the framework\\ to handle complex, multifactorial\\ scenarios.\end{tabular} \\ \hline
\end{tabular}}
\end{table}

\subsection{Baseline Models}
In our study, we selected three baseline models for comparison, each epitomizing state-of-the-art approaches in multi-agent forecasting, graph neural networks, and traffic prediction. These models were chosen based on their innovative methodologies and proven effectiveness in areas closely aligned with our research objectives.

\begin{itemize}
    \item \textbf{ELMA Method}~\cite{li2022elma}: Developed by Li et al.~\cite{li2022elma}, the ELMA method utilizes graph neural networks for forecasting multi-agent activities, particularly adept at handling spatiotemporal data. Its novelty lies in the use of energy-based learning, making it an excellent benchmark against our framework, which similarly leverages graph-based techniques in complex environments.
    \item \textbf{Self-Supervised Technique}~\cite{ma2021multi}: This technique is pioneering in self-supervised learning for predicting multi-agent driving behavior. Its relevance to our study comes from its focus on behavior prediction in diverse scenarios, using self-supervised domain knowledge—an advanced trend in multi-agent learning.
    \item \textbf{Internet Traffic Prediction Study}~\cite{jiang2022internet}: It involves internet traffic prediction using distributed multi-agent learning, employing LSTM and GRU models. GRU's superior performance in their study provides a valuable point of comparison for our research, which focuses on sophisticated learning techniques in traffic prediction.
\end{itemize}

Each of these models represents a significant stride in their respective fields. Their selection for comparison in our study is justified by their alignment with our research goals and their benchmark status in handling complex, dynamic datasets. By comparing our GraphRL framework against these models, we aim to demonstrate our approach's novelty and effectiveness in diverse real-world applications.

\subsection{Evaluation Metrics}

Mean Absolute Error (MAE) is a widely-utilized regression metric that gauges the average magnitude of errors between predicted and actual values in a dataset. It is calculated by averaging the absolute differences between these values, yielding a singular metric. Root Mean Squared Error (RMSE) is another prominent regression metric, assessing the average magnitude of differences between predicted and actual values. RMSE is computed as the square root of the mean of these squared differences. Mean Absolute Percentage Error (MAPE) represents yet another regression metric, quantifying the average absolute percentage error between predicted and actual values. It is derived by averaging the absolute differences between these values, expressed as a percentage of the actual values. Conversely, Cumulative Rewards is a performance metric specific to reinforcement learning. It measures the total rewards an agent accumulates over a specified timeframe or across a set number of actions, calculated by summing all rewards received during this period.

In the context of the experiments conducted for this study, Python version 3.7.6 served as the programming environment, with the deployment of several packages including TensorFlow, Keras, OpenAI Gym, and stable\_baselines3.

\begin{table}[]
\centering
\caption{Performance of the proposed framework in health forecasting}
\label{tab:health_monitor}
\begin{tabular}{@{}clcccc@{}}
\toprule
\multicolumn{1}{l}{} &
   &
  15 Min &
  30 Min &
  45 Min &
  60 Min \\ \midrule
\multicolumn{1}{c}{\multirow{3}{*}{\textbf{ELMA~\cite{li2022elma}}}} &
  \multicolumn{1}{l}{MAE} &
  \multicolumn{1}{c}{6.2} &
  \multicolumn{1}{c}{6.2} &
  \multicolumn{1}{c}{6.2} &
  \multicolumn{1}{c}{6.13} \\ \cmidrule(l){2-6} 
\multicolumn{1}{c}{} &
  \multicolumn{1}{l}{MAPE} &
  \multicolumn{1}{c}{13.91} &
  \multicolumn{1}{c}{13.91} &
  \multicolumn{1}{c}{13.91} &
  \multicolumn{1}{c}{13.91} \\ \cmidrule(l){2-6} 
\multicolumn{1}{c}{} &
  \multicolumn{1}{l}{RMSE} &
  \multicolumn{1}{c}{8.75} &
  \multicolumn{1}{c}{8.75} &
  \multicolumn{1}{c}{8.75} &
  \multicolumn{1}{c}{8.67} \\ \midrule
\multicolumn{1}{c}{\multirow{3}{*}{\textbf{GRU~\cite{ma2021multi}}}} &
  \multicolumn{1}{l}{MAE} &
  \multicolumn{1}{c}{0.95} &
  \multicolumn{1}{c}{0.95} &
  \multicolumn{1}{c}{0.97} &
  \multicolumn{1}{c}{0.98} \\ \cmidrule(l){2-6} 
\multicolumn{1}{c}{} &
  \multicolumn{1}{l}{MAPE} &
  \multicolumn{1}{c}{5.47} &
  \multicolumn{1}{c}{5.48} &
  \multicolumn{1}{c}{5.51} &
  \multicolumn{1}{c}{5.5} \\ \cmidrule(l){2-6} 
\multicolumn{1}{c}{} &
  \multicolumn{1}{l}{RMSE} &
  \multicolumn{1}{c}{1.25} &
  \multicolumn{1}{c}{1.25} &
  \multicolumn{1}{c}{1.27} &
  \multicolumn{1}{c}{1.28} \\ \midrule
\multicolumn{1}{c}{\multirow{3}{*}{\textbf{\begin{tabular}[c]{@{}c@{}}GRU-Based \\ \\ Multi-Agent~\cite{jiang2022internet}\end{tabular}}}} &
  \multicolumn{1}{l}{MAE} &
  \multicolumn{1}{c}{1.02} &
  \multicolumn{1}{c}{1.02} &
  \multicolumn{1}{c}{1.25} &
  \multicolumn{1}{c}{1.65} \\ \cmidrule(l){2-6} 
\multicolumn{1}{c}{} &
  \multicolumn{1}{l}{MAPE} &
  \multicolumn{1}{c}{8} &
  \multicolumn{1}{c}{3.47} &
  \multicolumn{1}{c}{4.53} &
  \multicolumn{1}{c}{5.27} \\ \cmidrule(l){2-6} 
\multicolumn{1}{c}{} &
  \multicolumn{1}{l}{RMSE} &
  \multicolumn{1}{c}{2.46} &
  \multicolumn{1}{c}{2.58} &
  \multicolumn{1}{c}{2.69} &
  \multicolumn{1}{c}{3.09} \\ \midrule
\multicolumn{1}{c}{\multirow{3}{*}{\textbf{GraphRL (Ours)}}} &
  \multicolumn{1}{l}{MAE} &
  \multicolumn{1}{c}{\textbf{0.56}} &
  \multicolumn{1}{c}{\textbf{0.87}} &
  \multicolumn{1}{c}{\textbf{0.68}} &
  \multicolumn{1}{c}{\textbf{0.7}} \\ \cmidrule(l){2-6} 
\multicolumn{1}{c}{} &
  \multicolumn{1}{l}{MAPE} &
  \multicolumn{1}{c}{\textbf{2.8}} &
  \multicolumn{1}{c}{\textbf{2.9}} &
  \multicolumn{1}{c}{\textbf{2.65}} &
  \multicolumn{1}{c}{\textbf{3.98}} \\ \cmidrule(l){2-6} 
\multicolumn{1}{c}{} &
  RMSE &
  \textbf{1.18} &
  \textbf{1.47} &
  \textbf{1.3} &
  \textbf{1.32} \\ \bottomrule
\end{tabular}
\end{table}
\section{Results and Analysis}\label{results}
In this section, the proposed framework performance in terms of time series forecasting and RL monitoring is compared to the baseline models in each application. 

\subsection{Predictive GraphRL Performance}
\textbf{Healthcare Forecasting:}
The proposed framework is evaluated to monitor health status by predicting future vital signs such as heart rate. Based on the sensor data and other clinical parameters such as ECG, Respiration, the time series prediction of the heart rate is conducted. The predicted values of heart for the next one hour are break-down into different time intervals (15 minutes, 30 minutes, 45 minutes, 60 minutes). Each of these time interval values acts as an observation for the GraphRL agent to monitor and communicate with the appropriate emergency team. The observation space of the vital sign, action space of different emergency teams and rewards for the agent actions in the predictive GraphRL environment are defined based on the modified early warning scores (MEWS)~\cite{signscanberra}. For the evaluation process, the WESAD dataset is adopted to conduct time series forecasting of heart rate. The proposed T-GCN in the predictive GraphRL environment performs better than the other baseline frameworks ELMA, GRU, and GRU-Based Multi-Agent as shown in Tab.~\ref{tab:health_monitor}. It achieves the lowest MAE, MAPE and RMSE values in all the time intervals.


\begin{table}[]
\caption{Performance of the proposed framework in traffic forecasting}
\label{tab:traffic_monitor}
\begin{tabular}{@{}clcccc@{}}
\toprule
\multicolumn{1}{l}{} &
   &
  15 Min &
  30 Min &
  45 Min &
  60 Min \\ \midrule
\multicolumn{1}{c}{\multirow{3}{*}{\textbf{ELMA~\cite{li2022elma}}}} &
  \multicolumn{1}{c}{MAE} &
  \multicolumn{1}{c}{6.73} &
  \multicolumn{1}{c}{6.73} &
  \multicolumn{1}{c}{6.73} &
  \multicolumn{1}{c}{6.72} \\ \cmidrule(l){2-6} 
\multicolumn{1}{c}{} &
  \multicolumn{1}{l}{MAPE} &
  \multicolumn{1}{c}{6.73} &
  \multicolumn{1}{c}{15.14} &
  \multicolumn{1}{c}{15.14} &
  \multicolumn{1}{c}{15.07} \\ \cmidrule(l){2-6} 
\multicolumn{1}{c}{} &
  \multicolumn{1}{l}{RMSE} &
  \multicolumn{1}{c}{6.72} &
  \multicolumn{1}{c}{9.4} &
  \multicolumn{1}{c}{9.4} &
  \multicolumn{1}{c}{9.39} \\ \midrule
\multicolumn{1}{c}{\multirow{3}{*}{\textbf{GRU~\cite{ma2021multi}}}} &
  \multicolumn{1}{c}{MAE} &
  \multicolumn{1}{c}{1.04} &
  \multicolumn{1}{c}{1.04} &
  \multicolumn{1}{c}{1.04} &
  \multicolumn{1}{c}{1.04} \\ \cmidrule(l){2-6} 
\multicolumn{1}{c}{} &
  \multicolumn{1}{l}{MAPE} &
  \multicolumn{1}{c}{6.04} &
  \multicolumn{1}{c}{6.01} &
  \multicolumn{1}{c}{5.96} &
  \multicolumn{1}{c}{6.1} \\ \cmidrule(l){2-6} 
\multicolumn{1}{c}{} &
  \multicolumn{1}{l}{RMSE} &
  \multicolumn{1}{c}{1.36} &
  \multicolumn{1}{c}{1.36} &
  \multicolumn{1}{c}{1.36} &
  \multicolumn{1}{c}{1.36} \\ \midrule
\multicolumn{1}{c}{\multirow{3}{*}{\textbf{\begin{tabular}[c]{@{}c@{}}GRU-Based \\ \\ Multi-Agent~\cite{jiang2022internet}\end{tabular}}}} &
  \multicolumn{1}{c}{MAE} &
  \multicolumn{1}{c}{1.85} &
  \multicolumn{1}{c}{1.85} &
  \multicolumn{1}{c}{1.96} &
  \multicolumn{1}{c}{1.82} \\ \cmidrule(l){2-6} 
\multicolumn{1}{c}{} &
  \multicolumn{1}{l}{MAPE} &
  \multicolumn{1}{c}{6.07} &
  \multicolumn{1}{c}{5.7} &
  \multicolumn{1}{c}{4.93} &
  \multicolumn{1}{c}{6.07} \\ \cmidrule(l){2-6} 
\multicolumn{1}{c}{} &
  \multicolumn{1}{l}{RMSE} &
  \multicolumn{1}{c}{2.88} &
  \multicolumn{1}{c}{3.21} &
  \multicolumn{1}{c}{3.43} &
  \multicolumn{1}{c}{3.54} \\ \midrule
\multicolumn{1}{c}{\multirow{3}{*}{\textbf{GraphRL (Ours)}}} &
  \multicolumn{1}{c}{MAE} &
  \multicolumn{1}{c}{\textbf{0.65}} &
  \multicolumn{1}{c}{\textbf{0.78}} &
  \multicolumn{1}{c}{\textbf{0.64}} &
  \multicolumn{1}{c}{\textbf{0.8}} \\ \cmidrule(l){2-6} 
\multicolumn{1}{c}{} &
  \multicolumn{1}{l}{MAPE} &
  \multicolumn{1}{c}{\textbf{4.1}} &
  \multicolumn{1}{c}{\textbf{7.85}} &
  \multicolumn{1}{c}{\textbf{5.65}} &
  \multicolumn{1}{c}{\textbf{7.99}} \\ \cmidrule(l){2-6} 
\multicolumn{1}{c}{} &
  RMSE &
  \textbf{1.27} &
  \textbf{1.22} &
  \textbf{1.26} &
  \textbf{1.41} \\ \bottomrule
\end{tabular}
\end{table}
\textbf{Traffic Forecasting:}
The goal of the proposed framework is to predict traffic using the predictive GraphRL environment. The system takes in data with the following features: EventId, Type, Severity, TMC, Description, StartTime, EndTime, TimeZone, LocationLat, LocationLng, Distance, AirportCode, Number, Street, Side, City, County, State, and ZipCode. The observation space includes the current traffic state, which is represented by the traffic events and their severity in a particular region. The actions referred to possible traffic management strategies, such as altering traffic light timings or changing the speed limit. The rewards are defined based on the efficiency of the chosen strategy, such as reduced travel time or decreased congestion. For all the baseline models and the proposed framework, the MAE, MAPE, and RMSE values are reported for forecasting at 15, 30, 45, and 60-minute intervals. As shown in Tab.~\ref{tab:traffic_monitor}, T-GCN outperforms the other models for all the forecasting intervals with the lowest MAE, MAPE, and RMSE values. The second-best performer is the GRU-Based Multi-Agent model, followed by GRU and ELMA.

\begin{table}[]
\centering
\caption{Performance of the proposed framework in weather forecasting}
\label{tab:weather_monitor}
\begin{tabular}{@{}clcccc@{}}
\toprule
\multicolumn{1}{l}{} &
   &
  15 Min &
  30 Min &
  45 Min &
  60 Min \\ \midrule
\multicolumn{1}{c}{\multirow{3}{*}{\textbf{ELMA~\cite{li2022elma}}}} &
  \multicolumn{1}{l}{MAE} &
  \multicolumn{1}{c}{6.69} &
  \multicolumn{1}{c}{6.69} &
  \multicolumn{1}{c}{6.69} &
  \multicolumn{1}{c}{6.65} \\ \cmidrule(l){2-6} 
\multicolumn{1}{c}{} &
  \multicolumn{1}{l}{MAPE} &
  \multicolumn{1}{c}{6.69} &
  \multicolumn{1}{c}{15.02} &
  \multicolumn{1}{c}{15.02} &
  \multicolumn{1}{c}{14.99} \\ \cmidrule(l){2-6} 
\multicolumn{1}{c}{} &
  \multicolumn{1}{l}{RMSE} &
  \multicolumn{1}{c}{6.69} &
  \multicolumn{1}{c}{9.39} &
  \multicolumn{1}{c}{9.39} &
  \multicolumn{1}{c}{9.34} \\ \midrule
\multicolumn{1}{c}{\multirow{3}{*}{\textbf{GRU~\cite{ma2021multi}}}} &
  \multicolumn{1}{l}{MAE} &
  \multicolumn{1}{c}{1.03} &
  \multicolumn{1}{c}{1.03} &
  \multicolumn{1}{c}{1.04} &
  \multicolumn{1}{c}{1.04} \\ \cmidrule(l){2-6} 
\multicolumn{1}{c}{} &
  \multicolumn{1}{l}{MAPE} &
  \multicolumn{1}{c}{5.96} &
  \multicolumn{1}{c}{5.94} &
  \multicolumn{1}{c}{5.93} &
  \multicolumn{1}{c}{6.02} \\ \cmidrule(l){2-6} 
\multicolumn{1}{c}{} &
  \multicolumn{1}{l}{RMSE} &
  \multicolumn{1}{c}{1.36} &
  \multicolumn{1}{c}{1.35} &
  \multicolumn{1}{c}{1.36} &
  \multicolumn{1}{c}{1.36} \\ \midrule
\multicolumn{1}{c}{\multirow{3}{*}{\textbf{\begin{tabular}[c]{@{}c@{}}GRU-Based \\ \\ Multi-Agent~\cite{jiang2022internet}\end{tabular}}}} &
  \multicolumn{1}{l}{MAE} &
  \multicolumn{1}{c}{1.65} &
  \multicolumn{1}{c}{1.65} &
  \multicolumn{1}{c}{1.85} &
  \multicolumn{1}{c}{2.02} \\ \cmidrule(l){2-6} 
\multicolumn{1}{c}{} &
  \multicolumn{1}{l}{MAPE} &
  \multicolumn{1}{c}{7.32} &
  \multicolumn{1}{c}{4.71} &
  \multicolumn{1}{c}{4.89} &
  \multicolumn{1}{c}{5.86} \\ \cmidrule(l){2-6} 
\multicolumn{1}{c}{} &
  \multicolumn{1}{l}{RMSE} &
  \multicolumn{1}{c}{2.76} &
  \multicolumn{1}{c}{2.99} &
  \multicolumn{1}{c}{3.16} &
  \multicolumn{1}{c}{3.43} \\ \midrule
\multicolumn{1}{c}{\multirow{3}{*}{\textbf{GraphRL (Ours)}}} &
  \multicolumn{1}{l}{MAE} &
  \multicolumn{1}{c}{\textbf{0.61}} &
  \multicolumn{1}{c}{\textbf{0.83}} &
  \multicolumn{1}{c}{\textbf{0.66}} &
  \multicolumn{1}{c}{\textbf{0.75}} \\ \cmidrule(l){2-6} 
\multicolumn{1}{c}{} &
  \multicolumn{1}{l}{MAPE} &
  \multicolumn{1}{c}{\textbf{3.95}} &
  \multicolumn{1}{c}{\textbf{5.88}} &
  \multicolumn{1}{c}{\textbf{5.15}} &
  \multicolumn{1}{c}{\textbf{7.99}} \\ \cmidrule(l){2-6} 
\multicolumn{1}{c}{} &
  RMSE &
  \textbf{1.23} &
  \textbf{1.12} &
  \textbf{1.28} &
  \textbf{1.26} \\ \bottomrule
\end{tabular}
\end{table}

\textbf{Weather Forecasting:}
In weather forecasting, the goal of the proposed framework is to use past weather data to predict future weather events and to optimise actions based on those predictions. In the predictive environment, the observation space is configured based on both the traffic and weather events datasets, including the event type, severity, start time, end time, location (latitude and longitude), and timezone. The actions represent the decisions the RL agent can take based on the observation space. For example, the agent could decide to issue a warning or alert for severe weather, adjust traffic signals or road signs, or change the speed limit on certain roads.  The agent could receive a reward for correctly predicting severe weather and issuing a timely warning. Using the proposed GraphRL framework allows modelling the relationships between different weather events and their impact on traffic in a more efficient way than traditional machine learning methods. The GraphRL agent learns from these relationships to make better decisions and improve its predictions over time. Comparing the different models, T-GCN had the best performance across all metrics and different time intervals: 15, 30, 45, and 60 minutes, followed by GRU-Based Multi-Agent, GRU, and ELMA. The results show that the T-GCN model had the lowest MAE, MAPE, and RMSE values for all forecasting intervals, indicating its superior forecasting performance compared to the other models as shown in Tab.~\ref{tab:weather_monitor}.

\subsection{GraphRL Agent Performance}
\begin{table}[]
\scriptsize
\caption{Proposed GraphRL Performance}
\label{tab:Agents}
\resizebox{\columnwidth}{!}{%
\begin{tabular}{@{}lrrr@{}}
\toprule
\textbf{AI Agents} &
  \multicolumn{1}{c}{\textbf{WESAD}} &
  \multicolumn{1}{c}{\textbf{\begin{tabular}[c]{@{}c@{}}LAM Traffic \\ Forecasting\end{tabular}}} &
  \multicolumn{1}{c}{\textbf{\begin{tabular}[c]{@{}c@{}}US Weather \\ Forecasting\end{tabular}}} \\ \midrule
Q Learning        & 43130          & 28840          & 39480          \\ \midrule
PPO               & 39480          & 33945          & 29480          \\ \midrule
A2C               & 41195          & 22845          & 40615          \\ \midrule
Double DQN        & 42615          & 25600          & 33945          \\ \midrule
DDPG              & 44600          & 34590          & 39945          \\ \midrule
DQN               & 41986          & 35219          & 40985          \\ \midrule
\textbf{GraphRL} & \textbf{48790} & \textbf{36195} & \textbf{53145} \\ \bottomrule
\end{tabular}}
\end{table}

The proposed RL agent was enabled with T-GCN and its performance is compared with other traditional RL agents as shown in Tab.~\ref{tab:Agents}. The table provides a comparison of different AI agents and their performance on three different datasets: WESAD, LAM Traffic Forecasting, and US Weather Forecasting. The performance of each agent is measured by a score, which is the total score achieved by the agent on the task over ten episodes. From the table, it can be seen that the proposed GraphRL agent is the most efficient agent on the WESAD dataset, as it scored the highest score. The DDPG and Q-Learning agents have the second-highest score on the WESAD dataset. On the LAM Traffic Forecasting dataset, the Q-Learning agent scored the lowest, and the proposed GraphRL agent scored the highest. On the US Weather Forecasting dataset, the A2C agent scored the lowest, while the GraphRL agent scored the highest. The GraphRL agent has outperformed other RL agents in all three predictive and monitoring applications.


\begin{figure}[h]
\begin{tikzpicture}
\begin{axis}[    xlabel={Episodes},    ylabel={Score},    legend pos=south east, legend style={font=\scriptsize},    yticklabel style={/pgf/number format/fixed},    ymin=-20000, ymax=65000,    width=\columnwidth, height=4.5cm,    xtick={1,2,3,4,5,6,7,8,9,10},    xticklabels={1,2,3,4,5,6,7,8,9,10}]
 
\addplot[color=blue, mark=*] coordinates {
    (1, 38925)
    (2, 43595)
    (3, 46875)
    (4, 57280)
    (5, 32245)
    (6, 485)
    (7, 48035)
    (8, 48790)
    (9, 42885)
    (10, 54470)
};

\addplot[color=red, mark=square*] coordinates {
    (1, -8810)
    (2, -8100)
    (3, -11845)
    (4, 31960)
    (5, 46295)
    (6, 32560)
    (7, 53640)
    (8, 36195)
    (9, 36480)
    (10, 46425)
};

\addplot[color=green, mark=triangle*] coordinates {
    (1, 23780)
    (2, -6765)
    (3, 33480)
    (4, 43635)
    (5, 24925)
    (6, 2670)
    (7, 8990)
    (8, 53145)
    (9, 58530)
    (10, 41405)
};
 
\legend{Health, Traffic, Weather}
 
\end{axis}
\end{tikzpicture}
    \caption{GraphRL Agent Rewards Distribution}
    \label{fig:GraphRL}
\end{figure}
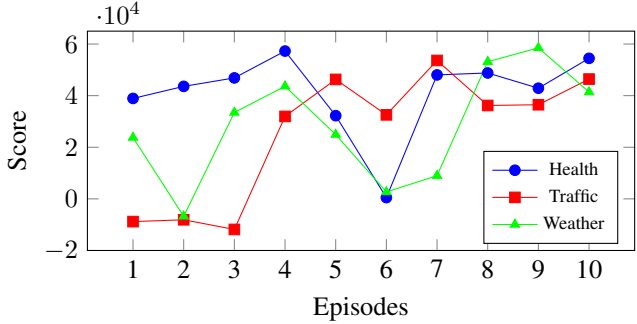
The performance of the GraphRL agent is measured by the episode score, which appears to be the total score achieved by the agent after a certain number of episodes. The breakdown of the proposed agent's score in each episode of the three applications is presented and compared in Fig.~\ref{fig:GraphRL}. The agent's performance on the WESAD dataset is relatively consistent, with the scores fluctuating between 32245 and 57280. On the LAM Traffic Forecasting dataset, the agent's performance is relatively inconsistent, with the scores fluctuating between -11845 and 46295. On the US Weather Forecasting dataset, the agent's performance is also relatively inconsistent, with the scores fluctuating between -6765 and 58530. This inconsistency of the scores is due to the exploration rate where the algorithm tries exploring all the actions randomly instead of using T-GCN model predictions.

\section{Bayesian Optimisation Results}\label{optimization}

\begin{figure}[h]
\begin{tikzpicture}
\begin{axis}[    xlabel={Episode},    ylabel={Rewards},    legend pos=north west, legend style={font=\scriptsize},   
legend cell align=left, ymajorgrids=true,    grid style=dashed,    ymin=-15000,    ymax=60000,    scaled y ticks = false,    y tick label style={/pgf/number format/sci},    width=\columnwidth,    height=4.5cm]

\addplot[color=red,mark=x] coordinates {
    (1, 537)
    (2, 1315)
    (3, 2990)
    (4, 2208)
    (5, 1011)
    (6, 1898)
    (7, -1878)
    (8, -5601)
    (9, -8802)
    (10, -13386)
};
\addlegendentry{$\alpha=0.1$}

\addplot[color=blue,mark=x] coordinates {
    (1, -2453)
    (2, -282)
    (3, 2428)
    (4, 1286)
    (5, 1465)
    (6, -692)
    (7, 6017)
    (8, 3712)
    (9, 3664)
    (10, 13624)
};
\addlegendentry{$\alpha=0.01$}

\addplot[color=green,mark=x] coordinates {
    (1, 1238)
    (2, 2994)
    (3, 5169)
    (4, 8232)
    (5, 12456)
    (6, 18356)
    (7, 26830)
    (8, 35428)
    (9, 45035)
    (10, 56545)
};
\addlegendentry{$\alpha=0.001$}

\addplot[color=black,mark=x] coordinates {
    (1, 494)
    (2, 520)
    (3, -1198)
    (4, -762)
    (5, -181)
    (6, -512)
    (7, -284)
    (8, 144)
    (9, -1175)
    (10, -2479)
};
\addlegendentry{$\alpha=0.0001$}

\addplot[color=purple,mark=x] coordinates {
    (1, -348)
    (2, -580)
    (3, -1244)
    (4, -820)
    (5, 113)
    (6, -542)
    (7, -2131)
    (8, -2882)
    (9, -4225)
    (10, -5163)
};
\addlegendentry{$\alpha=0.00001$}

\end{axis}
\end{tikzpicture}
\caption{Bayesian optimisation of $\alpha$ for GraphRL Agent}
\label{fig:alpha}
\end{figure}
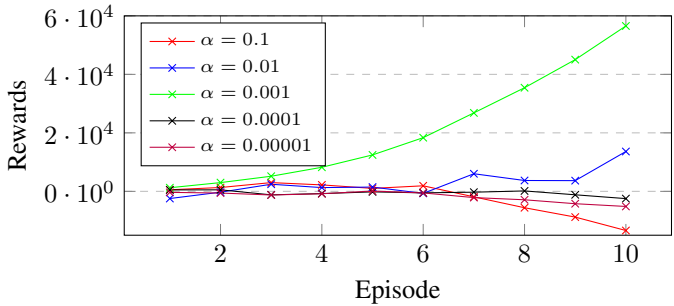


\begin{figure}[h]
\begin{tikzpicture}
\begin{axis}[    xlabel={Episode},    ylabel={Rewards},    legend pos=north west, legend style={font=\scriptsize},   
legend cell align=left, ymajorgrids=true,    grid style=dashed,    ymin=-15000,    ymax=60000,    scaled y ticks = false,    y tick label style={/pgf/number format/sci},    width=\columnwidth,    height=4.5cm]

\addplot[color=red,mark=x] coordinates {
    (1, -944)
    (2, -667)
    (3, 919)
    (4, 4009)
    (5, 7860)
    (6, 14303)
    (7, 21405)
    (8, 28799)
    (9, 38095)
    (10, 48823)
};

\addplot[color=blue,mark=x] coordinates {
    (1, -588)
    (2, 2080)
    (3, 3122)
    (4, 6573)
    (5, 9759)
    (6, 16085)
    (7, 25136)
    (8, 34720)
    (9, 46093)
    (10, 57388)
};

\addplot[color=green,mark=x] coordinates {
    (1, 909)
    (2, -258)
    (3, 1398)
    (4, 5207)
    (5, 10434)
    (6, 17728)
    (7, 25584)
    (8, 34898)
    (9, 32468)
    (10, 42637)
};

\addplot[color=orange,mark=x] coordinates {
    (1, -84)
    (2, 272)
    (3, 2355)
    (4, 5792)
    (5, 11162)
    (6, 17760)
    (7, 23002)
    (8, 31702)
    (9, 40771)
    (10, 52126)
};

\addplot[color=purple,mark=x] coordinates {
    (1, -389)
    (2, 799)
    (3, 4194)
    (4, 7551)
    (5, 11174)
    (6, 17898)
    (7, 24723)
    (8, 32023)
    (9, 40640)
    (10, 50340)
};
\legend{$\gamma=0.1$,$\gamma=0.01$,$\gamma=0.001$,$\gamma=0.0001$,$\gamma=0.00001$}
\end{axis}
\end{tikzpicture}
\caption{Bayesian optimisation of $\gamma$ for GraphRL Agent}
\label{fig:gamma}
\end{figure}
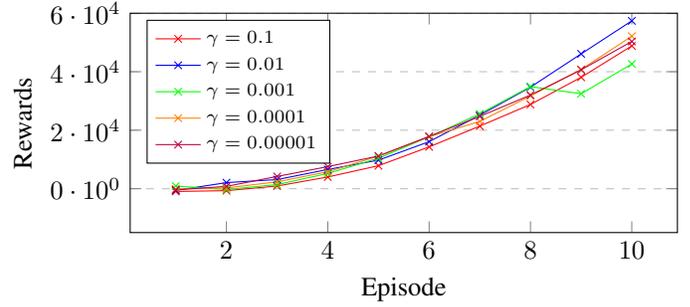

The results of Bayesian optimisation for the proposed GraphRL agent using different values of the learning rate parameter, $\alpha$, during different episodes are shown in Fig.~\ref{fig:alpha}. The values in the y-axis of the line chart represent the scores or rewards obtained by the agent during each episode. It can be observed that the performance of the agent varies for different values of alpha. For example, in episode 1, the agent performs better with alpha = 0.001 (20630) compared to the other values. Similarly, in episode 10, the agent performs better with alpha = 0.001 (97380) compared to the other values. These results suggest that the optimal value of alpha for the agent is $\alpha = 0.001$, and Bayesian optimisation can be used to find the best value of alpha for a given task.

These results in Fig.~\ref{fig:gamma} show the performance of an RL agent using temporal GCN for Q function approximation, using different values of the discount factor gamma. As we can see, the performance varies greatly depending on the value of the gamma chosen. A high value of gamma (0.95) results in poor performance, while lower values (0.75) result in better performance. This suggests that a lower discount factor is more appropriate, as it gives more weight to immediate rewards and less to future rewards. It also suggests that there is an optimal value of gamma, which would need to be further explored through more extensive experimentation.

\section{Conclusion}\label{conclusion}
The GraphRL framework, introduced in this study, embodies an innovative amalgamation of T-GCN and RL. It is specifically engineered to augment the prediction of future states in dynamic environments. Rigorous evaluations, utilizing an array of datasets such as WESAD, LA Traffic Forecasting, and US Weather Forecasting, have substantiated the framework's enhanced performance compared to conventional RL agents. Nonetheless, it is imperative to acknowledge that the efficacy of GraphRL is significantly contingent upon the caliber of the input data and necessitates substantial computational resources. The framework's reliance on data of high quality and structure constitutes a considerable limitation, with its accuracy and effectiveness being closely bound to the data's integrity. Additionally, the computational requisites, predominantly due to the T-GCN model integration, present challenges in scalability and broader applicability.

Future enhancements of the GraphRL framework will be directed towards surmounting these constraints and broadening its functional scope. Prospective developments entail the incorporation of spatial data processing, aimed at bolstering the framework's analytical prowess, particularly in processing data with spatial or geographical dimensions. Investigating a spectrum of graph-based models could yield insights for enhancing both the efficiency and efficacy of the framework. Furthermore, the exploration of real-time adaptive learning strategies presents a promising avenue for subsequent research. Such advancements are anticipated to enable the framework to dynamically adapt to evolving data patterns and environmental shifts. In summation, the GraphRL framework signifies a substantial advancement in the domain of time-series prediction and monitoring. Its adeptness in managing complex temporal data surpasses traditional RL methodologies, heralding innovative applications in sectors such as healthcare, traffic management, and environmental forecasting. As the framework undergoes continued refinement and evolution, it is positioned to emerge as an instrumental component in the progression of predictive analytics and intelligent monitoring systems, with extensive applicability across diverse fields.




\bibliographystyle{ieeetr}
\bibliography{bare_jrnl_new_sample4}

\vfill

\end{document}